\documentclass[10pt, a4paper, logo]{googledeepmind}
\usepackage{times}

\usepackage{hyperref}
\usepackage[leftcaption]{sidecap} % caption 在右边

\usepackage{amsmath,amsfonts,bm}
\usepackage{natbib}
\usepackage{fancyhdr}
\def\eqref#1{equation~\ref{#1}}
\def\1{\bm{1}}

\DeclareMathAlphabet{\mathsfit}{\encodingdefault}{\sfdefault}{m}{sl}
\SetMathAlphabet{\mathsfit}{bold}{\encodingdefault}{\sfdefault}{bx}{n}

\usepackage{hyperref}
\usepackage{url}
\usepackage{booktabs}
\usepackage{graphicx}
\usepackage{subcaption}
\usepackage{amssymb}
\usepackage[most]{tcolorbox}
\usepackage[table]{xcolor}
\tcbuselibrary{breakable} % 加载分页支持库
\usepackage{amsmath}
\usepackage{siunitx}
\usepackage{colortbl}

\usepackage{microtype}
\usepackage{color}
\usepackage{wrapfig}
\usepackage{natbib}
\usepackage{colortbl}
\usepackage{microtype}
\usepackage{graphicx}
\usepackage{booktabs} % for professional tables
\usepackage{array}
\usepackage{textcomp}
\usepackage{stfloats}
\usepackage{float}
\usepackage{verbatim}
\usepackage[ruled, linesnumbered, lined]{algorithm2e}
\usepackage{multirow}
\usepackage{enumitem}

\definecolor{BestColor}{HTML}{C8E6C9}  % 一个柔和的绿色
\definecolor{SecondBestColor}{HTML}{FFF9C4} % 一个非常淡的黄色

\usepackage{tcolorbox}
\usepackage{amsmath,amsfonts}

\definecolor{ggg}{RGB}{26,179,0}
\definecolor{rrr}{RGB}{179,0,0}
\definecolor{oodc}{RGB}{31,73,121}
\definecolor{idc}{RGB}{68,142,68}

\definecolor{mygray}{gray}{0.9}
\definecolor{lightblue}{RGB}{239,247,255}

\newcommand{\ie}{\emph{i.e., }}

\def\Bias#1#2{\bm{b}}
\newtcolorbox{examplebox}[2][]{ % 允许传入可选参数 [#1] 和必选标题参数 {#2}
    breakable, % 关键：允许跨页分割
    enhanced, % 增强模式（可选，支持更多样式）
    colback=white, % 框体内背景色
    colframe=cyan, % 边框颜色
    coltitle=white, % 标题文字颜色
    fonttitle=\bfseries, % 标题字体加粗
    title=#2, % 框体标题（第二个必选参数）
    overlay middle={\draw[cyan, line width=1pt](frame.south west)--(frame.south east);}, % 分割处添加横线
    overlay last={\draw[cyan, line width=1pt](frame.south west)--(frame.south east);}, % 最后一页底部横线
    #1 % 允许在调用时传入其他可选参数以覆盖默认样式
}

\usepackage[T1]{fontenc}
\usepackage{booktabs}      % 用于漂亮的表格线 (toprule, midrule, etc.)
\usepackage{graphicx}      % 用于 \resizebox
\usepackage[table]{xcolor} % 用于颜色
\usepackage{siunitx}       % 用于 S 列，实现小数点对齐
\usepackage{etoolbox}      % 用于 \ifstrequal，实现条件判断
\usepackage[normalem]{ulem}     % 用于更灵活的下划线 \uline

\definecolor{impcolor}{HTML}{2E8B57} % 提升使用的海绿色 (SeaGreen)

\newcommand{\improvementstyle}[1]{$^{\textcolor{impcolor}{\tiny #1}}$}

\newcommand{\scoreimp}[2]{%
  \textbf{#1}%
  \ifstrequal{#2}{+0.0}{}{%
    \ifstrequal{#2}{0.0}{}{%
      \makebox[0pt][l]{\improvementstyle{#2}}%
    }%
  }%
}

\title{PolicyAlign: Direct Policy-Based Safety Alignment for Large Language Models}

\author[1]{Chang Wu}
\author[2]{Junfeng Fang}
\author[3]{Houcheng Jiang}
\author[1]{Kai Tang}
\author[1]{Pengyu Cheng}
\author[1]{Xiaoxi Jiang}
\author[1]{Guanjun Jiang}
\author[1]{Xiang Wang}
\affil[1]{Qwen Large Model Application Team, Alibaba}
\affil[2]{National University of Singapore}
\affil[3]{Zhongguancun Academy}

\begin{abstract}
Safety alignment of large language models (LLMs) typically depends on high-quality supervision data, such as safe demonstrations or preference pairs. However, in real-world deployment, emerging safety requirements are often specified as natural-language policies, while corresponding supervision data may be costly, delayed, or unavailable. This creates a mismatch between rapidly evolving safety policies and conventional data-driven alignment methods. To address this, we propose \textbf{PolicyAlign}, a simple yet effective framework for directly aligning LLMs with safety policies. Given a safety policy, PolicyAlign first synthesizes policy-violating instructions and then performs on-policy self-distillation to internalize policy-guided behavior. To improve training stability and data efficiency, we further introduce \textbf{Policy-Sensitive Filtering}, which selects instructions where the policy induces the largest behavioral shift. Experiments across multiple models show that PolicyAlign consistently improves safety while maintaining low over-refusal and preserving general capabilities. PolicyAlign also generalizes to medical, legal, and financial safety scenarios, highlighting its potential as a scalable and maintainable approach to policy-based LLM safety alignment. The code is released at \url{https://github.com/Qwen-Applications/PolicyAlign}.
\end{abstract}

\begin{document}
\maketitle
%\vspace{-1mm}
\section{Introduction}
\label{sec:Introduction}

As large language models (LLMs) are increasingly deployed in open-ended settings, ensuring their safety remains an open problem \citep{wang2025comprehensive,lu2025alignment}, especially in high-stakes domains such as medicine, law, and finance \citep{aljohani2025comprehensive,cheng2025uncovering,chen2024survey}.
Most existing alignment methods depend on high-quality safety data. Supervised fine-tuning (SFT) \citep{wei2021finetuned} improves safety by training on human-curated safe demonstrations, while RL-based alignment methods \citep{ouyang2022training,bai2022training} further refine model behavior by optimizing against reward signals learned from human or model feedback. Despite their differences, these methods share a common requirement: safety requirements must be instantiated as curated demonstrations, preference annotations, or reward signals. Consequently, their effectiveness depends heavily on the quality and coverage of data. However, constructing such data is often time-consuming, labor-intensive, and in some cases requires collecting real-world unsafe examples \citep{pku-saferlhf,beavertails}.

\begin{figure}[t]
    \centering
    % \small
    \includegraphics[width=1.0\linewidth]{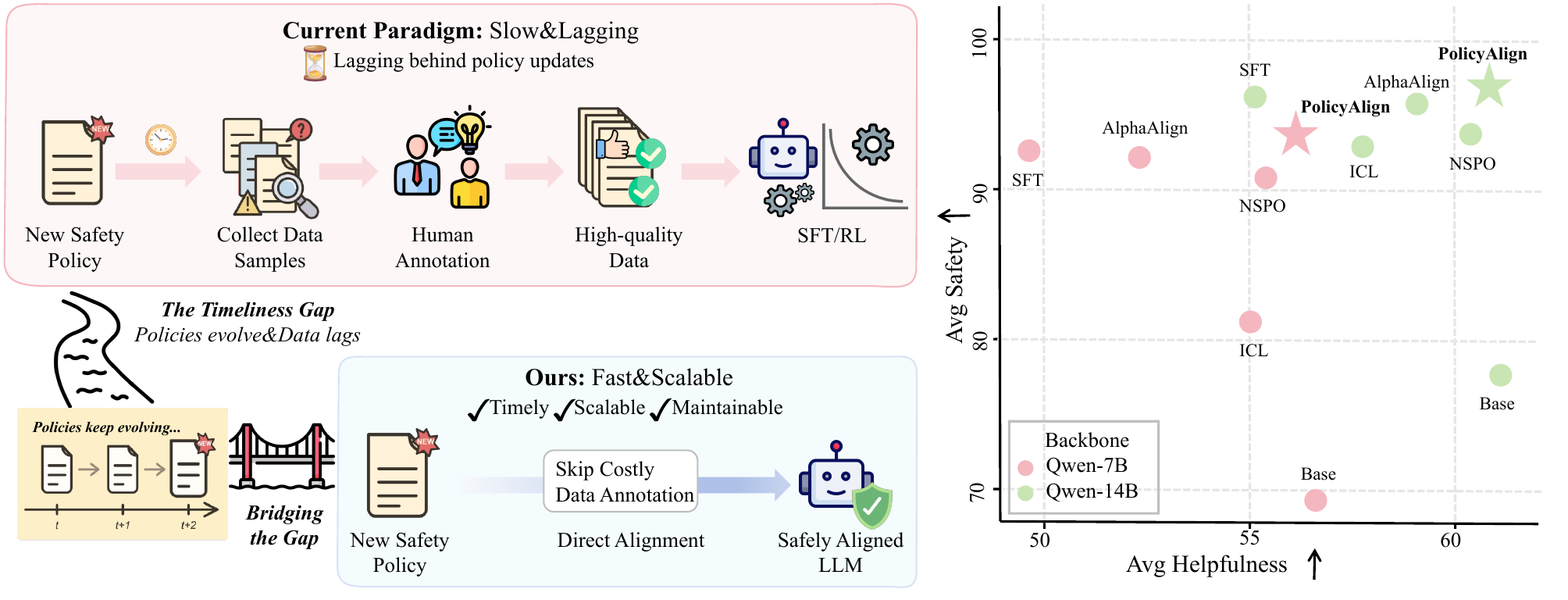}
    \vspace{-5mm}
    \caption{\textbf{Left:} Current alignment paradigm lags behind safety policy updates because new policies must first be converted into high-quality data before alignment. In contrast, our method directly aligns models with a new safety policy, avoiding expensive data annotation and enabling more timely, scalable, and maintainable alignment. \textbf{Right:} Trade-off between average helpfulness and safety on Qwen2.5-7B-Instruct and Qwen2.5-14B-Instruct. PolicyAlign achieves a better safety–helpfulness frontier than existing baselines, consistently improving safety while preserving helpfulness.}
     \label{fig:intro}
     \vspace{-5mm}
\end{figure}

This creates a fundamental mismatch: existing safety alignment methods require substantial effort to construct high-quality data, whereas real-world safety requirements are often specified as policies and demand rapid adaptation. In real-world deployment, emerging safety requirements are usually expressed in high-level policy documents, such as risk management frameworks, usage guidelines, and legal regulations \citep{nistAIRMF2023,whoAIHealth2021,euAIAct2024}. Such policies are widely available across domains, continuously evolving, and often released before high-quality data can be collected. Yet data-centric alignment methods cannot directly operate on such policies. Converting policies into effective training signals introduces substantial adaptation latency, 
especially for RL-based alignment pipelines that often involve multi-stage training. Rapidly adapting deployed LLMs to new policies also creates a dilemma: training only on new data risks catastrophic forgetting, while full retraining is too expensive.
As a result, existing alignment methods struggle to keep pace with evolving safety requirements (Figure~\ref{fig:intro}). 
For example, after the rapid rise of OpenClaw \citep{openclaw}, safety concerns quickly arose \citep{shan2026don}, leading governments to issue targeted safety policies on an urgent basis \citep{cncertopenclawrisk,openclawwarning}. Yet high-quality supervision data for these new requirements remains scarce. Bridging this gap would make LLM safety alignment more timely, scalable, and maintainable.

Existing attempts to use natural language principles (\ie safety white papers) as an alignment interface suffer from notable limitations. One approach directly prepends safety policies to the model context at inference time \citep{wei2026jailbreak}. While this can improve surface-level compliance, the desired safety behavior is not internalized into the model parameters. As policies become longer and more complex, such in-context learning (ICL) methods become increasingly unstable and fail to ensure faithful compliance. They also remain vulnerable to adversarial jailbreaks \citep{zou2023universal,andriushchenko2024jailbreaking}. Another approach converts policies into supervised examples and then trains the model with SFT \citep{bai2022constitutional}. Although this internalizes policy-guided behavior, it reintroduces the cost and latency of constructing high-quality data. These limitations point to the core challenge: how can LLMs efficiently and reliably internalize policy-guided safety behavior?

To explore the feasibility of aligning LLMs directly with safety policies, we propose \textbf{PolicyAlign}, a simple yet effective framework for policy-based safety alignment. Given a safety policy, PolicyAlign first uses a strong LLM to synthesize policy-relevant training scenarios, and then performs on-policy self-distillation \citep{minillm,agarwal2024policy,ye2026policy,lu2025onpolicy} to internalize the policy. By aligning models directly with policies rather than fitting large collections of curated safety input-output pairs, this paradigm enables more targeted adaptation. As a result, it helps mitigate overfitting and reduces the alignment tax \citep{huang2025safety}, namely the degradation of general capabilities such as knowledge understanding and reasoning.

In practice, however, we observe that synthesized instructions can vary substantially in quality and usefulness since they are not human-calibrated and often involve emerging risks that even strong commercial models may handle unreliably. This makes a straightforward application of this paradigm unstable.
To address this, we further introduce a \underline{P}olicy-\underline{S}ensitive \underline{F}iltering (\textbf{PSF}) mechanism that draws on curriculum learning \citep{curriculum}. PSF quantifies how much the policy changes the model's behavior for each sample and uses this signal to dynamically select the most appropriate instructions for the current model state. More broadly, PolicyAlign is a modular framework that is naturally compatible with advances in on-policy distillation and curriculum learning, allowing future improvements to be incorporated for more precise and efficient policy-based alignment.

We evaluate PolicyAlign across multiple models and safety scenarios to validate its effectiveness. Our experiments span models of different scales and architectures, including LLaMA-3.2-3B-Instruct \citep{llama3_2}, Qwen2.5-7B-Instruct, and Qwen2.5-14B-Instruct \citep{Qwen25}. The evaluation covers both general safety benchmarks, such as StrongREJECT \citep{strongreject} and AdvBench \citep{zou2023universal} for harmful requests, WildJailbreak \citep{wildteaming2024} and Fortress \citep{knight2025fortress} for adversarial attacks, and XSTest \citep{xstest} for over-refusal. To assess whether the framework generalizes to emerging safety domains, we further evaluate it across medical, legal, and financial safety settings. As shown in Figure~\ref{fig:intro}, compared with other alignment methods, PolicyAlign achieves the strongest overall safety performance while maintaining low over-refusal and preserving general capabilities. Together, these results provide strong evidence for the feasibility of aligning LLMs directly with safety policies. In summary, this work highlights the need for policy-based alignment in the era of rapidly evolving LLM safety requirements, and positions PolicyAlign as a simple and effective framework that opens a promising new path for LLM safety alignment.

\section{Preliminary}
\label{sec:preliminary}

\subsection{Safety Alignment from Policies}
\label{sec:policy_alignment_prelim}

Safety alignment aims to ensure that an LLM satisfies human-specified safety requirements while preserving general helpfulness. Existing alignment methods typically instantiate such requirements through high-quality training data, such as safe demonstrations for SFT \citep{wei2021finetuned} or reward signals from human or model for RL \citep{ouyang2022training,bai2022training}. These methods rely on training examples that explicitly encode desirable behavior for specific instructions. 
However, in many real-world safety scenarios, the most readily available form of supervision is not input-output pairs but a natural-language safety policy, such as a usage guideline, risk management framework, or domain-specific regulation \citep{nistAIRMF2023,whoAIHealth2021,euAIAct2024}.

We formalize policy-based safety alignment as follows.
Let $\mathcal{C}$ denote a safety policy that specifies high-level behavioral constraints, and let $\mathcal{X}$ denote the target instruction space.
For each instruction $x \in \mathcal{X}$, the policy $\mathcal{C}$ implicitly defines a set of safe responses under $\mathcal{C}$, denoted by $\mathcal{S}_{\mathcal{C}}(x)$. A response $y \notin \mathcal{S}_{\mathcal{C}}(x)$ is considered a policy violation.
Let $\mathcal{X}_{\mathrm{harmful}} \subseteq \mathcal{X}$ and $\mathcal{X}_{\mathrm{benign}} \subseteq \mathcal{X}$ denote harmful and benign instructions, respectively.
For $x \in \mathcal{X}_{\mathrm{harmful}}$, $\mathcal{S}_{\mathcal{C}}(x)$ typically includes responses that refuse direct harmful assistance, avoid operational details, and when appropriate, provide safe alternatives. 
For $x \in \mathcal{X}_{\mathrm{benign}}$, the model should remain useful and avoid unnecessary refusal.
We therefore define $\mathcal{H}_{\mathcal{C}}(x) \subseteq \mathcal{S}_{\mathcal{C}}(x)$ as the set of responses that are both safe and helpful for benign instructions.

A model aligned with $\mathcal{C}$ should satisfy two requirements: it should be safe on harmful instructions and helpful on benign instructions.
Formally, this can be expressed as:
\begin{align}
    \Pr_{y \sim \pi_{\theta}(\cdot \mid x)}
    \left[
    y \in \mathcal{S}_{\mathcal{C}}(x)
    \right]
    &\geq 1-\epsilon,
    \qquad x \in \mathcal{X}_{\mathrm{harmful}},
    \label{eq:safety_constraint}
    \\
    \Pr_{y \sim \pi_{\theta}(\cdot \mid x)}
    \left[
    y \in \mathcal{H}_{\mathcal{C}}(x)
    \right]
    &\geq 1-\delta,
    \qquad x \in \mathcal{X}_{\mathrm{benign}}.
    \label{eq:helpfulness_constraint}
\end{align}
Here, $\epsilon > 0$ denotes a small tolerated probability of policy violation, and $\delta > 0$ denotes a small tolerated probability of unhelpful or unnecessarily restrictive behavior.

The main challenge is that $\mathcal{S}_{\mathcal{C}}(x)$ and $\mathcal{H}_{\mathcal{C}}(x)$ cannot be directly observed, and manually curating them is costly, especially for new or domain-specific safety policies.
PolicyAlign addresses this challenge by using the policy document $\mathcal{C}$ itself as supervision.
Instead of converting $\mathcal{C}$ into safe demonstrations or preference pairs, we use it to induce a policy-conditioned teacher distribution, from which a policy-unconditioned student learns aligned behavior.

\subsection{Problem Setup}
\label{sec:setup}

We formulate policy-based alignment as transferring behavior from a policy-conditioned distribution to a policy-unconditioned model. Given an instruction $x$, an LLM parameterized by $\theta$ generates a response $y=(y_1,\ldots,y_L)$ autoregressively:
\begin{equation}
    \pi_{\theta}(y \mid x)
    =
    \prod_{t=1}^{L}
    \pi_{\theta}(y_t \mid x, y_{<t}).
\end{equation}
Our goal is to train a safety-aligned model by internalizing $\mathcal{C}$ into the model parameters, so that the model can generate policy-consistent responses without requiring $\mathcal{C}$ at inference time.

To this end, we instantiate two roles from the same underlying model architecture.
The \emph{student} is a policy-unconditioned model $\pi_{\theta}(y \mid x)$ that receives only the user instruction.
The \emph{teacher} is a policy-conditioned model $\pi_{\mathrm{ref}}(y \mid x, \mathcal{C})$ that receives both the instruction $x$ and the safety policy $\mathcal{C}$ as input.
The objective is to transfer the policy-induced behavior of the teacher to the student, thereby enabling policy-consistent generation at inference time.

\section{Method}
\label{sec:method}

\begin{figure}[t]
    \centering
    \includegraphics[width=0.87\linewidth]{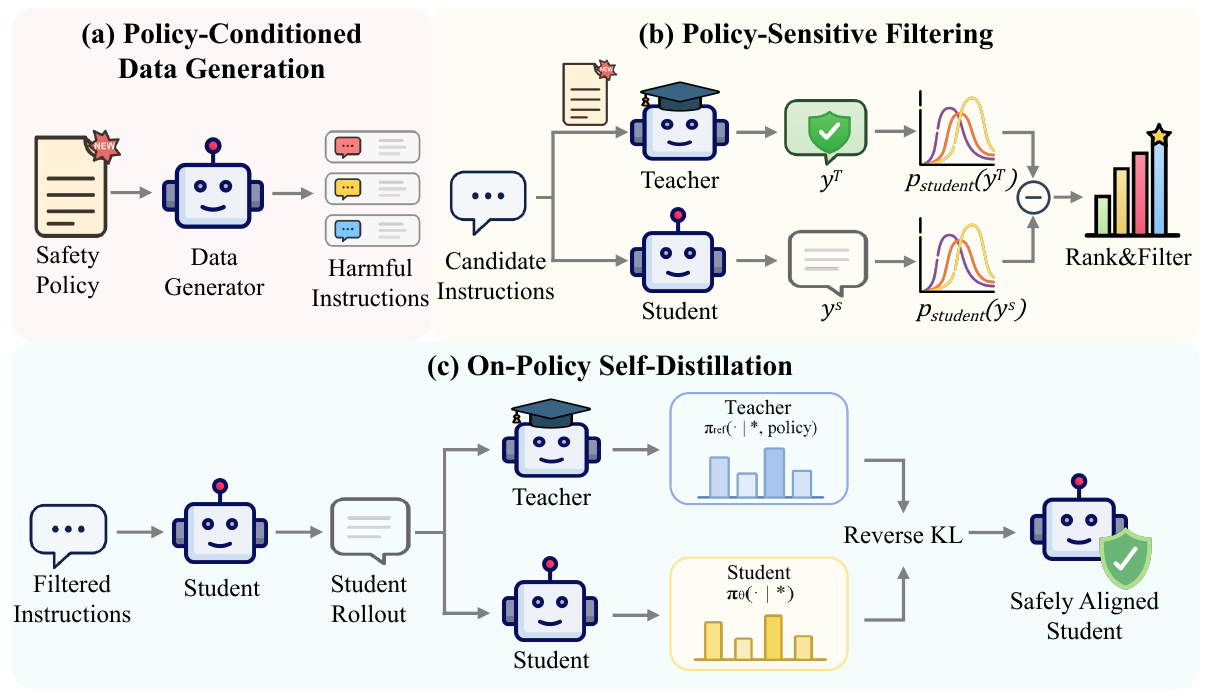}
    % \vspace{-2mm}
    \caption{Overview of PolicyAlign. (a) Policy-conditioned data generation synthesizes harmful instructions from a given safety policy. (b) Policy-Sensitive Filtering ranks candidate instructions by the behavioral gap between policy-conditioned teacher responses and policy-unconditioned student responses. (c) On-policy self-distillation trains the student on its own trajectories and minimizes token-level reverse KL loss between the student and the policy-conditioned teacher distributions.}
    \label{fig:framework}
     \vspace{-3mm}
\end{figure}

In this section, we introduce \textbf{PolicyAlign}, which directly aligns models with safety policies. Specifically, given a policy, we first use a strong LLM to synthesize a set of harmful instructions conditioned on the safety policy (Section~\ref{sec:data_gen}). We then score each instruction by how much the policy changes the model's behavior and select the most appropriate instructions for training (Section~\ref{sec:PSF}). Finally, we train the model to internalize the safety policy through on-policy self-distillation (Section~\ref{sec:self-opd}). An overview of the framework is shown in Figure~\ref{fig:framework}.

\subsection{Policy-Conditioned Data Generation}
\label{sec:data_gen}

Aligning a model with a safety policy requires training on policy-violating instructions, since these are the scenarios in which the policy should change the model's behavior.
Manually curating such instructions is costly, especially when the policy targets emerging risks for which representative harmful examples are scarce.

We automate this step by prompting a strong LLM $\mathcal{G}$ (\ie a frontier model such as GPT-5.4) with the safety policy $\mathcal{C}$ to synthesize a diverse set of $N$ harmful instructions that violate the policy:
\begin{equation}
\mathcal{D}=\{x_i\}_{i=1}^{N}, \quad x_i \sim \mathcal{G}(\cdot \mid \mathcal{C}).
\label{eq:data_gen}
\end{equation}
The generation prompt instructs $\mathcal{G}$ to cover multiple violation categories, severity levels, and linguistic variations to span the scope of the safety policy.
Importantly, this process requires only the safety policy itself, without human annotation or existing safety datasets.

The resulting instructions serve as candidate training scenarios. However, because they are LLM-generated, their quality can be uneven, particularly in emerging or specialized domains where $\mathcal{G}$ may have limited coverage. Some samples may be noisy, trivial, or poorly matched to the current student. This motivates the targeted filtering mechanism described next.

\subsection{Policy-Sensitive Filtering}
\label{sec:PSF}

We observe that not all synthesized instructions are equally useful for alignment. Instructions that the model already handles safely provide little training signal, while poorly constructed samples can introduce noise.
To address the inconsistent quality of synthetic data and improve data efficiency, we propose \textbf{Policy-Sensitive Filtering (PSF)}, a mechanism inspired by curriculum learning \citep{curriculum} that selects instructions for which the safety policy leads to the largest behavioral shift in the model.

\textbf{Scoring.} For each candidate instruction $x_i$, we generate two responses:
\begin{itemize}[leftmargin=*,nosep]
\item Teacher response: $y_T\sim\pi_{\mathrm{ref}}(\cdot\mid x_i,\mathcal{C})$\,, the policy-guided safe response.
\item Student response: $y_S\sim\pi_\theta(\cdot\mid x_i)$\,, the model's default response without the safety policy.
\end{itemize}
We evaluate both responses under the student $\pi_\theta(\cdot\mid x_i)$ and compute the \emph{policy sensitivity score}:
\begin{equation}
s(x_i)\;=\;\bar{\ell}_\theta(y_S\mid x_i)\;-\;\bar{\ell}_\theta(y_T\mid x_i),
\label{eq:psf_score}
\end{equation}
where $\bar{\ell}_\theta(y\mid x)=\frac{1}{L}\sum_{t=1}^{L}\log\pi_\theta(y_t\mid y_{<t},x)$ denotes the length-normalized log-likelihood under the student, with $L = |y|$.

\textbf{Interpretation.}
The score $s(x_i)$ quantifies the degree to which the student prefers its own response over the teacher's response.
A large $s(x_i)$ indicates a significant behavioral gap: the student confidently produces its default (potentially unsafe) response but assigns low probability to the teacher's policy-guided (safe) response.
These are precisely the instructions where training will yield the most meaningful behavioral change.
Conversely, a small $s(x_i)$ indicates that the model already behaves similarly with or without the policy, making the instruction less informative for alignment.

\textbf{Selection.}
We rank all candidate instructions by their policy sensitivity scores and retain the top-$k$ instructions:
\begin{equation}
\mathcal{D}_{\mathrm{filtered}}=\mathrm{Top}\text{-}k\!\big(\{x_i\}_{i=1}^N,\;s\big).
\label{eq:selection}
\end{equation}
By focusing training on the most policy-sensitive instructions, PSF improves data efficiency and filters out noisy or uninformative samples, concentrating learning on instructions where the policy provides the strongest corrective signal.

\subsection{On-Policy Self-Distillation}
\label{sec:self-opd}

With the filtered instruction set $\mathcal{D}_{\mathrm{filtered}}$, we train the student to internalize the teacher's safety behavior via on-policy self-distillation. The key idea is to let the student sample responses from its own current distribution and then align its next-token distribution with the policy-conditioned teacher. This transfers policy-compliant behavior into the student parameters.

\textbf{On-policy rollout.}
For each instruction $x\in\mathcal{D}_{\mathrm{filtered}}$, the student generates a complete response by sampling from its current distribution:
\begin{equation}
\hat{y}=(\hat{y}_1,\dots,\hat{y}_L)\sim\pi_\theta(\cdot\mid x).
\label{eq:rollout}
\end{equation}
Training on student-generated samples is critical: it ensures the learning signal reflects the student's actual generation distribution, avoiding the exposure bias that arises when training on static or teacher-generated sequences \citep{minillm,agarwal2024policy}.

\textbf{Reverse KL objective.}
We optimize the student by minimizing the reverse KL divergence to the policy-conditioned teacher on student-generated prefixes:
\begin{equation}
\mathcal{L}(\theta)
=\mathbb{E}_{x\sim\mathcal{D}_{\mathrm{filtered}},\,  \hat{y}\sim\pi_\theta(\cdot\mid x)}
\left[
\sum_{t=1}^{L}
D_{\mathrm{KL}}\!\left(
\pi_\theta(\cdot\mid \hat{y}_{<t},x)
\;\middle\|\;
\pi_{\mathrm{ref}}(\cdot\mid \hat{y}_{<t},x,\mathcal{C})
\right)
\right].
\label{eq:loss}
\end{equation}

\textbf{Why reverse KL?} 
For safety alignment, the key objective is not to cover every plausible teacher response, but to prevent the student from assigning probability mass to policy-inconsistent behaviors. Thus, we naturally adopt the reverse KL divergence. Compared to the forward KL, reverse KL encourages the student to concentrate on the teacher's high-confidence behaviors and avoid weakly supported regions that may correspond to unsafe or policy-inconsistent responses.

\section{Experiments}
\label{sec:exp}

\begin{table*}[t]
\centering
\scriptsize
\setlength{\tabcolsep}{1.0pt}
\setlength{\arrayrulewidth}{0.35pt}
\renewcommand{\arraystretch}{1.0}
\caption{Main results on safety, over-refusal, and utility benchmarks. The best results are \textbf{bold}.}
\resizebox{\textwidth}{!}{%
\begin{tabular}{l|cc|cc|c|c|ccc|c}
\toprule
\multirow{2}{*}{Model} 
& \multicolumn{2}{|c|}{Harmful ($\downarrow$)} 
& \multicolumn{2}{c|}{Jailbreak ($\downarrow$)} 
& \multicolumn{1}{c|}{Safety Avg. ($\downarrow$)}
& \multicolumn{1}{c|}{XSTest ($\uparrow$)} 
& \multicolumn{3}{c|}{Utility ($\uparrow$)}
& \multicolumn{1}{c}{Utility Avg. ($\uparrow$)} \\
\cmidrule(lr){2-3} 
\cmidrule(lr){4-5} 
\cmidrule(lr){6-6} 
\cmidrule(lr){7-7} 
\cmidrule(lr){8-10} 
\cmidrule(lr){11-11}
& StrongReject & AdvBench 
& WildJailbreak & Fortress 
& Avg. 
& XSTest 
& MMLU-Pro & GPQA & MATH 
& Avg. \\
\midrule

\multicolumn{11}{c}{LLaMA-3.2-3B-Instruct} \\
\midrule
Base         & 4.15 & 2.12 & 38.60 & 32.4 & 19.32 & \textbf{92.0} & 31.07 & 23.74 & 19.4 & 24.74 \\
ICL          & \textbf{0.32} & 0.19 & 22.71 & 9.4 & 8.16 & 68.8 & \textbf{32.21} & 22.22 & \textbf{21.0} & \textbf{25.14} \\
SFT          & 0.64 & 2.12 & \textbf{9.50} & 10.2 & 5.62 & 73.2 & 28.22 & 21.72 & 13.8 & 21.25 \\
AlphaAlign   & 0.64 & 1.92 & 11.63 & 10.0 & 6.05 & 82.4 & 29.88 & 22.22 & 18.0 & 23.37 \\
NSPO         & 0.96 & 2.12 & 11.90 & 10.4 & 6.35 & 83.6 & 30.87 & 23.23 & 18.8 & 24.30 \\
GRPO+Policy  & 0.96 & \textbf{0} & 12.26 & 9.2 & 5.61 & 79.4 & 31.27 & 23.23 & 19.8 & 24.77 \\
\rowcolor{lightblue}
\textbf{PolicyAlign} & \textbf{0.32} & \textbf{0} & 11.18 & \textbf{7.4} & \textbf{4.73} & 90.0 & 31.43 & \textbf{24.24} & 18.8 & 24.82 \\
\midrule

\multicolumn{11}{c}{Qwen2.5-7B-Instruct} \\
\midrule
Base         & 2.24 & 0.58 & 62.26 & 61.0 & 31.52 & \textbf{94.4} & \textbf{51.70} & 26.26 & 53.4 & 43.79 \\
ICL          & 0.32 & 0.38 & 33.26 & 43.2 & 19.29 & 85.2 & 51.61 & \textbf{28.78} & \textbf{54.2} & \textbf{44.86} \\
SFT          & 0.64 & \textbf{0} & 13.67 & 17.8 & 8.03 & 81.6 & 45.95 & 27.78 & 40.4 & 38.04 \\
AlphaAlign   & \textbf{0} & 0.19 & 13.57 & 18.2 & 7.99 & 84.0 & 48.37 & 27.27 & 47.6 & 41.08 \\
NSPO         & 0.32 & 0.19 & 13.67 & 18.4 & 8.15 & 84.8 & 50.86 & 27.27 & 52.6 & 43.58 \\
GRPO+Policy  & \textbf{0} & 0.38 & 13.22 & 16.4 & 7.50 & 82.0 & 50.69 & 27.78 & 52.4 & 43.62 \\
\rowcolor{lightblue}
\textbf{PolicyAlign} & \textbf{0} & \textbf{0} & \textbf{12.40} & \textbf{15.4} & \textbf{6.95} & 88.8 & 51.25 & 28.28 & \textbf{54.2} & 44.58 \\
\midrule

\multicolumn{11}{c}{Qwen2.5-14B-Instruct} \\
\midrule
Base         & 1.28 & 0.19 & 47.83 & 46.4 & 23.93 & \textbf{96.4} & 59.28 & 37.88 & 58.2 & 51.79 \\
ICL          & 0.64 & \textbf{0} & 12.59 & 16.6 & 7.46 & 82.4 & 58.52 & 33.84 & \textbf{58.4} & 50.25 \\
SFT          & 2.24 & 0.19 & \textbf{4.55} & 7.6 & 3.65 & 80.0 & 53.24 & 36.36 & 50.6 & 46.73 \\
AlphaAlign   & 0.32 & 0.19 & 4.75 & 9.4 & 3.67 & 86.8 & 58.07 & 35.86 & 54.4 & 49.44 \\
NSPO         & 0.64 & 0.38 & 4.89 & 9.8 & 3.93 & 87.6 & 58.74 & 36.87 & 57.4 & 51.00 \\
GRPO+Policy  & \textbf{0} & 0.38 & 4.98 & 8.6 & 3.49 & 86.2 & 58.44 & 37.37 & 57.2 & 51.00 \\
\rowcolor{lightblue}
\textbf{PolicyAlign} & \textbf{0} & \textbf{0} & 4.75 & \textbf{7.2} & \textbf{2.99} & 90.4 & \textbf{59.64} & \textbf{38.38} & 58.2 & \textbf{52.07} \\
\bottomrule
\end{tabular}%
}
\label{tab:main_results}
\vspace{-5pt}
\end{table*}

In this section, we conduct experiments to answer the following research questions:
\begin{itemize}[leftmargin=1em]
    \item \textbf{RQ1:} Can PolicyAlign effectively improve model safety while preserving general capabilities?
    \item \textbf{RQ2:} Can PolicyAlign generalize to emerging domain-specific safety scenarios where high-quality supervision data is scarce?
    \item \textbf{RQ3:} How important is Policy-Sensitive Filtering (PSF) to the effectiveness of PolicyAlign?
    \item \textbf{RQ4:} How does PolicyAlign scale with the amount of policy-based training data?
\end{itemize}

\subsection{Experimental Setup}

\textbf{Models.} We evaluate PolicyAlign on three instruction-tuned LLMs spanning different scales and architectures:
LLaMA-3.2-3B-Instruct \citep{llama3_2},
Qwen2.5-7B-Instruct, and Qwen2.5-14B-Instruct \citep{Qwen25}.

\textbf{Baselines.}
We compare PolicyAlign with the following baselines:
(1) Base, the original instruction-tuned model without additional safety alignment;
(2) ICL, which prepends the safety policy to the system prompt at inference time;
(3) SFT, which supervised fine-tunes the model on policy-based safe demonstrations generated by our data generator;
(4) AlphaAlign \citep{alphaalign}, an RL-based method that encourages proactive safety reasoning;
(5) NSPO \citep{nspo}, which preserves general capabilities through null-space constrained policy optimization; and
(6) GRPO+Policy, which applies Group Relative Policy Optimization \citep{deepseekmath} with a policy-conditioned reward model.

\textbf{Evaluation.} We evaluate PolicyAlign on five safety benchmarks and three general capability benchmarks. For safety, we use StrongREJECT \citep{strongreject} and AdvBench \citep{zou2023universal} for harmful requests, WildJailbreak \citep{wildteaming2024} and Fortress \citep{knight2025fortress} for jailbreak robustness, and XSTest \citep{xstest} for over-refusal. For the first four benchmarks, we use Qwen3Guard-Gen-8B \citep{qwen3guard} as the evaluator and report Attack Success Rate (ASR). For XSTest, we use GPT-4o \citep{gpt4o} as the evaluator and report the full Compliance Rate (CR). To assess whether safety alignment degrades general capabilities, we evaluate models on MMLU-Pro \citep{mmlu-pro} for knowledge understanding, GPQA-Diamond \citep{gpqa} for graduate-level scientific reasoning, and MATH500 \citep{math500} for math reasoning. We report Accuracy on all general capability benchmarks. 

\textbf{Implementation Details.}
For each backbone, the teacher and student are initialized from the same instruction-tuned checkpoint. The teacher receives the target safety policy as additional context, while the student is trained without policy input to internalize the policy-conditioned behavior. The teacher is updated as an exponential moving average (EMA) of the student with decay $0.99$. We optimize the student using token-level reverse KL divergence between the student distribution and the policy-conditioned EMA teacher distribution along student-generated trajectories. For efficiency, we approximate the KL term using the top-$k$ tokens under the student distribution, with $k=256$ in all experiments.
We train each model for 50 steps for LLM safety and 100 steps for financial, legal, and medical safety. We use a learning rate of $5 \times 10^{-6}$, a batch size of 128, and a maximum training response length of 512 tokens. During evaluation, all methods use the same decoding configuration within each benchmark: temperature 0 and a maximum generation length of 1024 tokens. Additional experimental details are provided in Appendix~\ref{app:experimental_setup}.

\subsection{Main Results (RQ1)}
\label{sec:main_results}

Table~\ref{tab:main_results} reports the main results on safety, over-refusal, and general capability benchmarks. We highlight three key observations:

\begin{itemize}[leftmargin=*]

\item \textbf{PolicyAlign achieves strong safety alignment.} 
PolicyAlign consistently improves safety across all three backbone models and all safety benchmarks.
On AdvBench, it reduces ASR to 0\% for all backbones, and also achieves the best performance on StrongReject.
The gains are particularly pronounced on the more challenging jailbreak benchmarks: PolicyAlign obtains the lowest ASR on Fortress for every backbone, reducing ASR over the base models by up to 25.0\% for LLaMA-3.2-3B-Instruct, 45.6\% for Qwen2.5-7B-Instruct, and 39.2\% for Qwen2.5-14B-Instruct.

\item \textbf{PolicyAlign mitigates over-refusal compared with existing alignment methods.}
A key distinction between PolicyAlign and other training-based alignment methods is its behavior on benign but superficially unsafe queries. On XSTest, existing training-based baselines largely reduce compliance. For example, on Qwen2.5-7B-Instruct, SFT decreases the compliance rate from 94.4\% to 81.6\% ($-12.8$), while AlphaAlign reduces it to 84.0\% ($-10.4$). In contrast, PolicyAlign preserves a higher compliance rate of 88.8\%, with the same trend observed across model scales.
These results suggest that PolicyAlign better captures the intent of the safety policy rather than simply increasing refusal rates, enabling more nuanced decisions on borderline queries.

\item \textbf{PolicyAlign preserves general capabilities while achieving a strong safety-utility trade-off.}
PolicyAlign maintains strong performance across general capability benchmarks while consistently improving safety. Compared with existing training-based alignment baselines, it better preserves the base model's general capabilities. For example, on Qwen2.5-7B-Instruct, SFT reduces MMLU-Pro from 51.70 to 45.95 and MATH500 from 53.4 to 40.4, whereas PolicyAlign maintains MMLU-Pro at 51.25 and improves MATH500 to 54.2.
A similar pattern holds for the other two backbones.
These results suggest that policy-conditioned self-distillation enables more targeted safety adaptation than fitting supervised demonstrations, thereby preserving the model's knowledge and reasoning abilities.

\end{itemize}

\begin{figure}[t]
\centering
\includegraphics[width=1.0\linewidth]{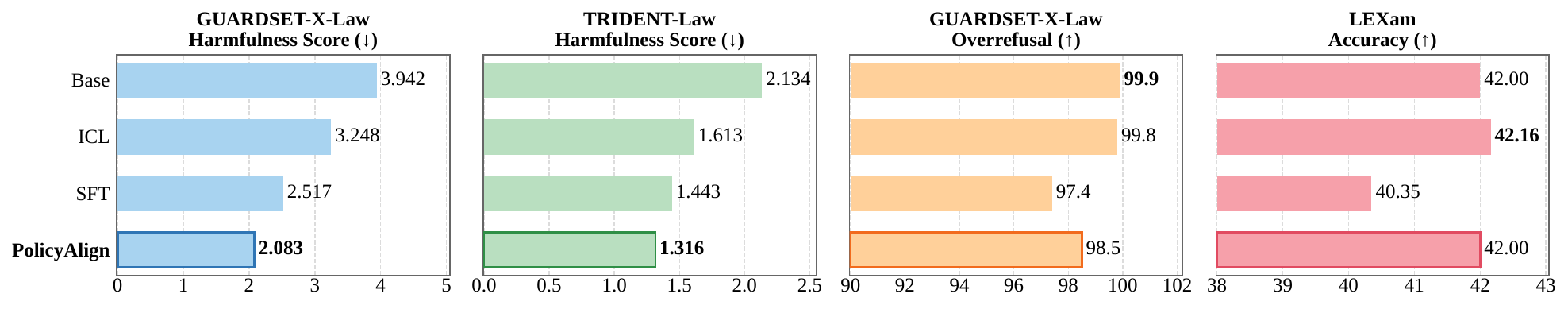}
\caption{Results on legal domain safety and general capability benchmarks.}
\label{fig:legal_results}
\vspace{-5pt}
\end{figure}

\begin{figure}[t]
\centering
\includegraphics[width=1.0\linewidth]{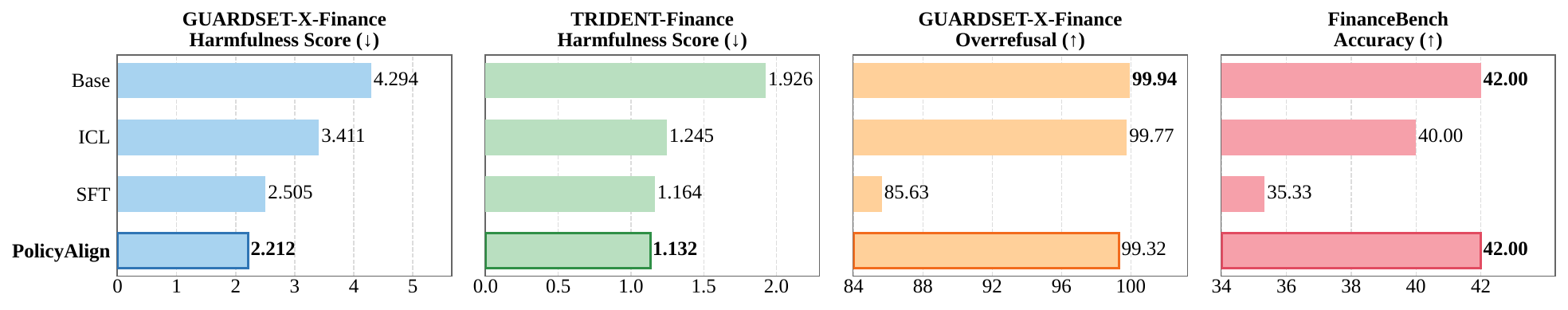}
\caption{Results on finance domain safety and general capability benchmarks.}
\label{fig:finance_results}
\vspace{-3pt}
\end{figure}

\begin{figure}[t]
\centering
\begin{minipage}[t]{0.35\textwidth}
\centering
\includegraphics[width=\linewidth]{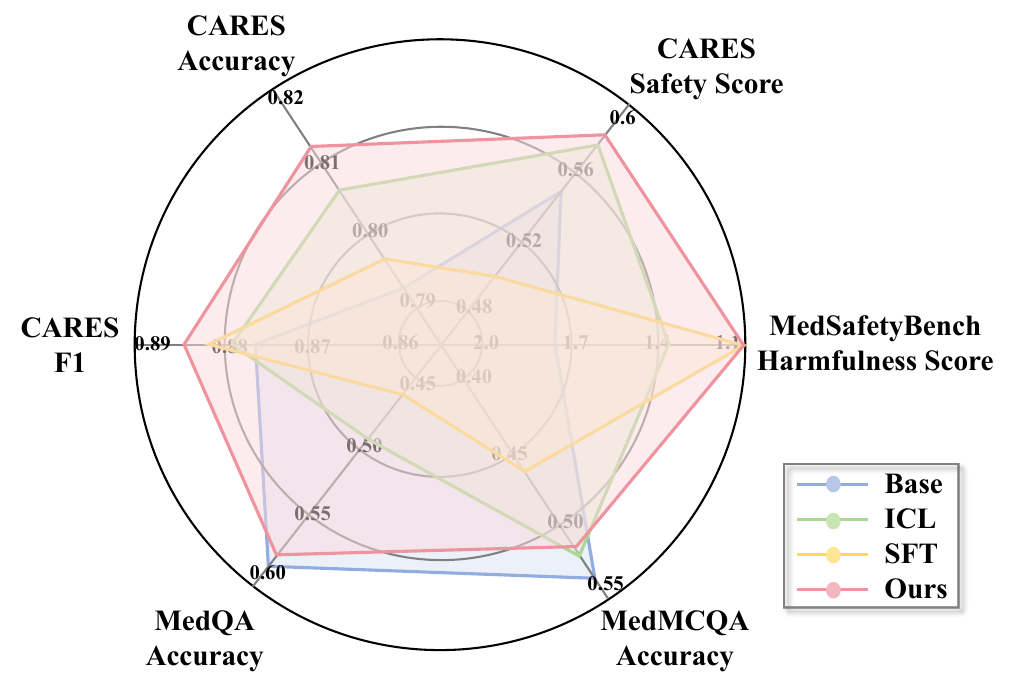}
\caption{Results on medical domain safety and general capability benchmarks.}
\label{fig:medical_results}
\end{minipage}
\hfill
\begin{minipage}[t]{0.3\textwidth}
\centering
\includegraphics[width=\linewidth]{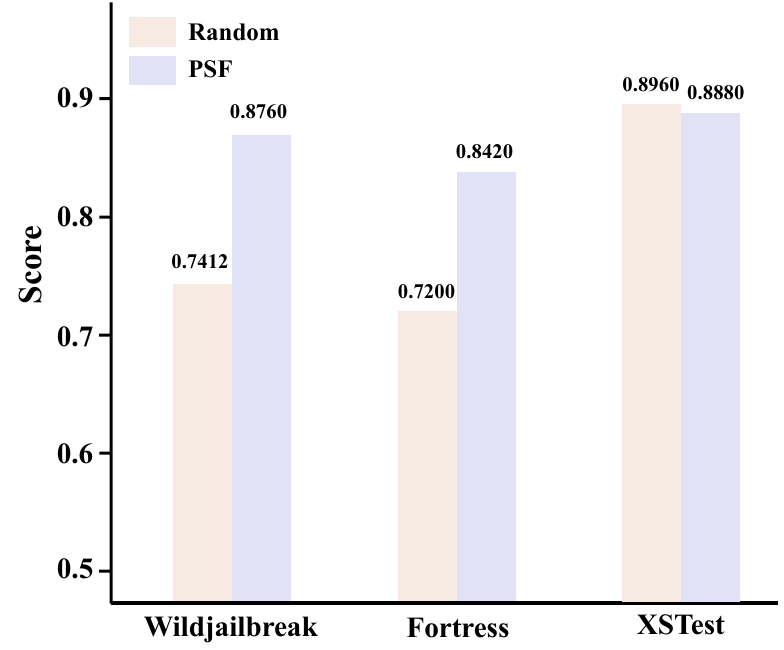}
\caption{Ablation on the PSF mechanism. }
\label{fig:ablation_psf}
\end{minipage}
\hfill
\begin{minipage}[t]{0.3\textwidth}
\centering
\includegraphics[width=\linewidth]{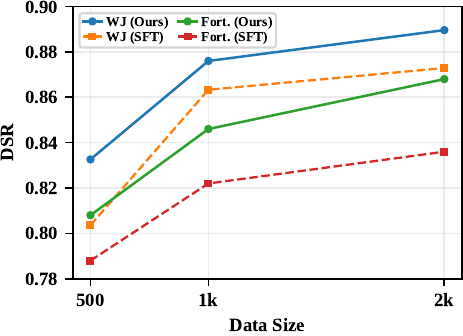}
\caption{Effect of training data scale.}
\label{fig:data_scaling}
\end{minipage}
\vspace{-5pt}
\end{figure}

\subsection{Generalization to Emerging Domain-Specific Safety Scenarios (RQ2)}
\label{sec:domain_results}

We next evaluate whether PolicyAlign generalizes to emerging high-stakes domains. This setting is central to our motivation: in these domains, safety requirements are often first specified as policies, guidelines, or regulations, while high-quality supervision data may be unavailable or expensive to collect. We conduct all domain-specific experiments on Qwen2.5-7B-Instruct and evaluate three domains: medicine, law, and finance.

\textbf{Evaluation.}
For medical safety, we evaluate on MedSafetyBench \citep{medsafetybench} and CARES \citep{cares}.
Following MedSafetyBench's original evaluation protocol, we prompt the model with each harmful request and use GPT-4o as the judge to assign a harmfulness score from 1 to 5 to the response and report the average harmfulness score.
For CARES, which evaluates medical safety under both benign and adversarially rewritten prompts, we use GPT-4o to classify each response as Accept, Caution, or Refuse, and report Safety Score, Accuracy, and F1. For medical capability, we evaluate on MedQA \citep{medqa} and MedMCQA \citep{medmcqa}, two medical multiple-choice QA benchmarks, and report Accuracy.
For legal and financial safety, we use the law and finance subsets of GUARDSET-X \citep{guardsetx} and TRIDENT \citep{trident}.
For unsafe queries, we use GPT-4o to score the harmfulness of model responses from 1 to 5 and report the average harmfulness score. For benign queries, we measure the compliance rate on safe requests as the over-refusal evaluation and report the full Compliance Rate.
For legal capability, we use the four-choice subset of LEXam \citep{lexam} and report Accuracy. For financial capability, we use FinanceBench \citep{financebench}, evaluating answer correctness with GPT-4o and reporting accuracy.

\textbf{Legal safety.} Figure~\ref{fig:legal_results} reports results in the legal domain. PolicyAlign achieves the lowest Harmfulness Score on both legal safety benchmarks. For example, on GUARDSET-X, it reduces harmfulness from 3.942 to 2.083, outperforming all baselines. At the same time, it maintains strong benign-query behavior, obtaining 98.5 Compliance Rate on over-refusal and 42.00 Accuracy on LEXam. Compared with SFT, PolicyAlign achieves stronger legal safety while avoiding degradation of legal capability.

\textbf{Financial safety.}
Figure~\ref{fig:finance_results} shows a similar trend in the financial domain. PolicyAlign reduces Harmfulness Score from 4.294 to 2.212 on GUARDSET-X and from 1.926 to 1.132 on TRIDENT, achieving the best safety performance among all methods. Crucially, it preserves financial capability: PolicyAlign maintains 42.00 on FinanceBench, matching the base model, whereas SFT drops to 35.33. SFT also causes a large decrease in the over-refusal score, from 99.94 to 85.63, while PolicyAlign retains a high score of 99.32.

\textbf{Medical safety.} Figure~\ref{fig:medical_results} shows that PolicyAlign improves medical safety while largely preserving medical capability. On MedSafetyBench, PolicyAlign reduces the Harmfulness Score from 1.753 to 1.100. Moreover, PolicyAlign achieves the best results on CARES across all three metrics. These results suggest that PolicyAlign improves safety beyond simple refusal behavior, extending to more nuanced medical safety classification and response generation. Meanwhile, PolicyAlign preserves medical capability much better than SFT. For example, on MedQA, SFT drops to 0.4595, whereas PolicyAlign reaches 0.5727, close to the base model's 0.5892.

\textbf{Across all three domains, PolicyAlign consistently improves safety while avoiding severe capability degradation and over-refusal.} These results validate that, in emerging domains where safety policies are available but high-quality data is scarce, PolicyAlign provides an effective and low-cost path to safety alignment while largely preserving domain-specific utility.

\subsection{Ablation Study on Policy-Sensitive Filtering (RQ3)}
\label{sec:ablation}

To verify the effectiveness of PSF, we compare it with random instruction selection under the same training budget. We conduct the ablation on Qwen2.5-7B-Instruct, with results shown in Figure~\ref{fig:ablation_psf}.

\begin{itemize} [leftmargin=*]
    \item \textbf{PSF provides the largest gains on the most challenging cases.} On the more challenging adversarial benchmarks, PSF substantially improves Defense Success Rate (DSR) from 0.7412 to 0.8760 on WildJailbreak and from 0.7200 to 0.8420 on Fortress. Meanwhile, XSTest compliance remains comparable, indicating that these safety gains do not come from increased refusal of safe queries. By concentrating a fixed training budget on instructions for which the policy most changes model behavior, PSF acts as a form of curriculum learning \citep{curriculum}, focusing on high-value examples and enabling more efficient, targeted alignment than random sampling.
\end{itemize}

\subsection{Effect of Training Data Scale (RQ4)}
\label{sec:data_scaling}

We study how the size of the training data affects performance. 
We train PolicyAlign and SFT with 500, 1k, and 2k policy-based examples, and report Defense Success Rate (DSR) on WildJailbreak and Fortress in Figure~\ref{fig:data_scaling}. We find that:

\begin{itemize}[leftmargin=*]

\item \textbf{PolicyAlign benefits from more data and consistently outperforms SFT.}
Increasing the number of training examples improves the DSR of both PolicyAlign and SFT on WildJailbreak and Fortress. For PolicyAlign, the improvement from 500 to 1k examples is larger than the gain from 1k to 2k examples, suggesting diminishing returns within the tested range. Nevertheless, PolicyAlign continues to improve as more data is added, showing that the framework can effectively scale with additional policy-based examples.
Across all data sizes, PolicyAlign achieves higher DSR than SFT on both benchmarks. These results show that PolicyAlign effectively leverages policy-based data while maintaining a clear advantage over supervised fine-tuning under different data budgets.

\end{itemize}

\section{Limitation \& Future Work}
\label{sec:limitations}

Although PolicyAlign demonstrates strong performance across general and domain-specific safety settings, our current evaluation focuses on single-turn, text-only scenarios. Extending PolicyAlign to multi-turn conversational jailbreaks and multimodal inputs remains an important direction for future work. In addition, our experiments assume that the target safety policy can be provided to the teacher in a clear and usable form. Real-world policies may be long, ambiguous, or partially overlapping, which may require policy decomposition, interpretation, and reasoning before distillation.

Looking forward, PolicyAlign offers a promising step toward policy-based safety alignment in dynamic real-world environments. As safety requirements and regulations continue to evolve, future research should investigate policy versioning, conflict resolution across overlapping guidelines, and continual alignment methods. These directions are important for maintaining policy-consistent models under evolving requirements, while avoiding the latency and cost of repeatedly reconstructing large-scale supervised alignment datasets or retraining models from scratch.

\section{Conclusion}
\label{sec:conclusion}

In this work, we introduce PolicyAlign that directly aligns LLMs with safety policies. By selecting the most valuable training instructions with Policy-Sensitive Filtering and distilling policy-conditioned behavior into the model through on-policy self-distillation, PolicyAlign internalizes safety requirements without relying on costly data annotation. Experiments across multiple backbones, safety and utility benchmarks, and high-stakes domains show that PolicyAlign improves safety while better preserving benign-query compliance and general capabilities. Overall, PolicyAlign offers a timely, scalable, and effective path for aligning LLMs with rapidly evolving real-world safety policies.

\bibliography{conference}
\bibliographystyle{conference}

\appendix
\clearpage

\section{Related Work}
\label{app:related_work}

\textbf{LLM Safety Alignment.}
LLM safety alignment aims to guide model behavior toward human-specified safety requirements. Recent advancements in LLM safety mainly rely on high-quality safety data.
Supervised fine-tuning (SFT) \citep{wei2021finetuned} trains models on human-curated safe demonstrations, while RL-based alignment methods \citep{ouyang2022training,bai2022training} further refine model behavior by optimizing against reward signals. Building on these paradigms, recent methods improve the training objective by introducing safety constraints into preference optimization \citep{safedpo}, using RL objectives for safety reasoning \citep{alphaalign}, or preserving general capabilities by null-space constrained policy optimization \citep{nspo}. Despite their effectiveness, these methods remain dependent on high-quality training data, which can be slow and costly to collect, especially when safety policies evolve rapidly or new risks emerge.
In contrast to these approaches, PolicyAlign directly aligns models with safety policies and enables rapid and scalable adaptation to emerging safety guidelines while avoiding the high costs of data annotation.

\textbf{On-Policy Distillation.} On-policy distillation (OPD) \citep{minillm, agarwal2024policy, ye2026policy, lu2025onpolicy} trains a student model by minimizing the KL divergence between the student and a teacher on student-generated trajectories. By optimizing on the states that the student actually encounters during generation, OPD mitigates the exposure bias caused by offline training on fixed teacher-generated responses. Recent studies have used OPD to internalize contextual or privileged information into model parameters.
For example, OPCD \citep{ye2026policy} internalizes system prompts, SDFT \citep{sdft} enables continual learning from demonstrations, and SDPO \citep{sdpo} distills rich textual feedback into policy updates. 
Despite these advances in context and feedback internalization, the potential of OPD for safety alignment remains underexplored.
We bridge this gap by formulating policy-based safety alignment as an on-policy self-distillation problem, where a policy-conditioned model serves as the teacher and a policy-unconditioned student learns to internalize policy-consistent behavior.

\section{Experimental Setup}
\label{app:experimental_setup}

Here we provide additional details on evaluation benchmarks,  computing resources, and safety policies across LLM safety, finance, law, and medicine.

\subsection{Benchmarks}
\label{app:benchmarks}

We evaluate PolicyAlign on safety, over-refusal, general capability, and domain-specific benchmarks. We introduce the benchmarks used in experiments.

\subsubsection{General Safety and Over-Refusal Benchmarks}
\label{app:general_safety_benchmarks}

\begin{itemize}[leftmargin=*]
    \item \textbf{StrongREJECT} \citep{strongreject} contains 313 high-quality forbidden prompts spanning six harmful-behavior categories, enabling evaluation of whether model responses provide concrete and useful assistance for disallowed requests.
    
    \item \textbf{AdvBench} \citep{zou2023universal} contains 500 harmful behaviors formulated as direct instructions. It is widely used to evaluate whether aligned LLMs can be induced to comply with malicious requests.
    
    \item \textbf{WildJailbreak} \citep{wildteaming2024} is derived from the WildTeaming framework, which mines jailbreak tactics from in-the-wild user-chatbot interactions. The benchmark includes both vanilla harmful requests and adversarial jailbreak prompts, together with benign look-alike queries.
    
    \item \textbf{Fortress} \citep{knight2025fortress} evaluates safeguard robustness against adversarial prompts in national security and public safety scenarios. Its public set covers high-risk domains such as criminal or financial illicit activities, with matched benign counterparts for over-refusal analysis.
    
    \item \textbf{XSTest} \citep{xstest} is designed to identify exaggerated safety behaviors in LLMs. It contains safe prompts that superficially resemble unsafe requests, along with unsafe contrast prompts, making it suitable for measuring whether models refuse benign user intents.
\end{itemize}

For StrongREJECT, AdvBench, WildJailbreak, and Fortress, we use Qwen3Guard-Gen-8B \citep{qwen3guard} as the evaluator and report ASR. For XSTest, we use GPT-4o \citep{gpt4o} as the evaluator and report the full Compliance Rate.

\subsubsection{General Capability Benchmarks}
\label{app:general_capability_benchmarks}

\begin{itemize}[leftmargin=*]
    \item \textbf{MMLU-Pro} \citep{mmlu-pro} is a more challenging and robust extension of MMLU. It introduces more reasoning-focused questions, expands the answer space from four to ten choices, and removes trivial or noisy questions from the original benchmark.
    
    \item \textbf{GPQA-Diamond} \citep{gpqa} is the highest-quality subset of GPQA, a graduate-level, Google-proof science QA benchmark. It contains expert-written multiple-choice questions in biology, physics, and chemistry that are difficult even for skilled non-experts with web access.
    
    \item \textbf{MATH500} \citep{math500} is a 500-problem held-out subset of the MATH benchmark. The problems are sampled from competition-style mathematics and are commonly used to evaluate multi-step mathematical reasoning and final-answer correctness.
\end{itemize}

We report Accuracy on all general capability benchmarks. These benchmarks are used to assess whether safety alignment degrades the model's general knowledge and reasoning ability.

\subsubsection{Domain-Specific Safety Benchmarks}
\label{app:domain_safety_benchmarks}

\begin{itemize}[leftmargin=*]
    \item \textbf{MedSafetyBench} \citep{medsafetybench} is a medical safety benchmark designed to evaluate whether LLMs produce harmful medical responses. It defines medical safety based on the American Medical Association Principles of Medical Ethics and focuses on risks to personal health, public health, patient safety, and human rights.
    
    \item \textbf{CARES} \citep{cares} evaluates medical LLM safety under clinically grounded adversarial and ambiguous conditions. It contains prompts spanning multiple medical safety principles, harm levels, and prompting styles, including direct, indirect, obfuscated, and role-play attacks.
    
    \item \textbf{GUARDSET-X} \citep{guardsetx} is a guardrail benchmark covering safety risks across multiple high-stakes domains. It includes both unsafe and benign queries, as well as attack-enhanced inputs, to evaluate whether models can identify domain-specific safety violations while avoiding over-refusal. In our experiments, we use its legal and finance subsets to assess domain-specific safety alignment.

    \item \textbf{TRIDENT} \citep{trident} is a domain-specific safety benchmark grounded in professional norms and ethical requirements. It evaluates whether LLMs avoid unsafe or inappropriate assistance in high-stakes domains while preserving useful responses to benign requests. In our experiments, we use its legal and finance subsets to evaluate safety alignment in law and finance.
\end{itemize}

For MedSafetyBench, we follow the original evaluation protocol and use GPT-4o to assign a harmfulness score from 1 to 5, reporting the average harmfulness score. For CARES, GPT-4o classifies each response as \textit{Accept}, \textit{Caution}, or \textit{Refuse}, and we report Safety Score, Accuracy, and F1.
For GUARDSET-X and TRIDENT in the legal and financial domains, GPT-4o scores harmful responses from 1 to 5, and we report the average harmfulness score; for benign queries, we report the full Compliance Rate to measure over-refusal.

\subsubsection{Domain-Specific Capability Benchmarks}
\label{app:domain_capability_benchmarks}

\begin{itemize}[leftmargin=*]
    \item \textbf{MedQA} \citep{medqa} is a medical question-answering benchmark collected from professional medical board exams.
    It consists of free-form multiple-choice medical questions and is commonly used to evaluate clinical knowledge and medical reasoning.
    
    \item \textbf{MedMCQA} \citep{medmcqa} is a large-scale multi-subject medical multiple-choice QA benchmark.
    It contains more than 194k questions from AIIMS and NEET-PG entrance exams, covering thousands of healthcare topics across 21 medical subjects.
    
    \item \textbf{LEXam} \citep{lexam} is a legal reasoning benchmark derived from 340 law exams across 116 law school courses.
    It includes both open-ended and multiple-choice legal questions in English and German; in our experiments, we use the four-choice English subset.
    
    \item \textbf{FinanceBench} \citep{financebench} is an open-book financial question-answering benchmark based on publicly traded companies. It contains finance questions with corresponding answers and evidence strings, and is designed to evaluate whether LLMs can answer realistic financial analysis questions with factual grounding.
\end{itemize}

For medical capability, we evaluate on MedQA and MedMCQA and report Accuracy.
For legal capability, we use the four-choice English subset of LEXam and report Accuracy. For financial capability, we evaluate FinanceBench using GPT-4o as the evaluator and report Accuracy.

\begin{table*}[t]
\centering
\scriptsize
\setlength{\tabcolsep}{1.2pt}
\setlength{\arrayrulewidth}{0.35pt}
\renewcommand{\arraystretch}{1.0}
\caption{Results on safety, over-refusal, and utility benchmarks for Qwen3-4B. The best results are \textbf{bold}.}
\label{tab:qwen3_4b}
\resizebox{\textwidth}{!}{%
\begin{tabular}{l|cc|cc|c|c|ccc|c}
\toprule
\multirow{2}{*}{Model}
& \multicolumn{2}{|c|}{Harmful ($\downarrow$)}
& \multicolumn{2}{c|}{Jailbreak ($\downarrow$)}
& \multicolumn{1}{c|}{Safety Avg. ($\downarrow$)}
& \multicolumn{1}{c|}{XSTest ($\uparrow$)}
& \multicolumn{3}{c|}{Utility ($\uparrow$)}
& \multicolumn{1}{c}{Utility Avg. ($\uparrow$)} \\
\cmidrule(lr){2-3}
\cmidrule(lr){4-5}
\cmidrule(lr){6-6}
\cmidrule(lr){7-7}
\cmidrule(lr){8-10}
\cmidrule(lr){11-11}
& StrongREJECT & AdvBench
& WildJailbreak & Fortress
& Avg.
& XSTest
& MMLU-Pro & GPQA & MATH
& Avg. \\
\midrule

\multicolumn{11}{c}{Qwen3-4B} \\
\midrule
Base & 7.35 & 2.31 & 69.46 & 37.2 & 29.08 & \textbf{94.8} & 66.29 & \textbf{50.00} & 72.8 & 63.03 \\
ICL & \textbf{0} & 0.38 & 21.09 & 24.2 & 11.42 & 71.2 & \textbf{68.41} & 47.47 & 74.0 & 63.29 \\
SFT & 1.60 & 0.77 & 9.73 & 17.0 & 7.28 & 67.2 & 65.88 & 47.47 & 70.8 & 61.38 \\
\rowcolor{lightblue}
\textbf{PolicyAlign} & 0.32 & \textbf{0} & \textbf{8.51} & \textbf{13.6} & \textbf{5.61} & 87.6 & 68.29 & 47.98 & \textbf{74.8} & \textbf{63.69} \\
\bottomrule
\end{tabular}%
}
\vspace{-5pt}
\end{table*}

\subsection{Computing Resources}
\label{app:compute_resources}

All experiments are conducted on NVIDIA A100-80GB GPUs. 
The average training time is approximately 10 GPU hours for 7B models and 16 GPU hours for 14B models.

\subsection{Safety Policies}
\label{app:safety_policies}

PolicyAlign is designed to align model behavior with explicit safety policies. In our experiments, we consider one general LLM safety policy and three domain-specific safety policies. We use authoritative policy documents as the normative basis for defining safe and unsafe model behavior. To make these policies directly usable by LLMs, we lightly rewrite and reorganize the original documents into system-prompt-style instructions. These rewritten policies preserve the core safety requirements of the source documents, while expressing them as actionable rules for model behavior.

\begin{itemize}[leftmargin=*]

    \item \textbf{General LLM safety policy.}
    For general-purpose safety alignment, we base our policy on the NIST AI Risk Management Framework: Generative AI Profile \citep{nist_ai}. 
    This policy covers broad generative-AI risks, including dangerous or violent content, CBRN-related assistance, information security, data privacy, harmful bias, abusive content, and information integrity. 
    We adapt these risk categories into a general LLM safety policy.

    \item \textbf{Medical safety policy.}
    For medicine, we base our policy on the American Medical Association Principles of Medical Ethics \citep{ama_principles}. 
    These principles emphasize competent and compassionate care, professional integrity, respect for the law, patient confidentiality and privacy, continued medical education, community health, patient welfare, and access to care.

    \item \textbf{Legal safety policy.}
    For law, we base our policy on the American Bar Association Model Rules of Professional Conduct~\citep{aba_model_rules}. 
    The policy covers core legal-professional obligations such as competence, diligence, communication, reasonable fees, confidentiality, conflicts of interest, duties to former or prospective clients, unauthorized practice, and professional integrity. 

    \item \textbf{Financial safety policy.}
    For finance, we base our policy on the CFA Institute Code of Ethics and Standards of Professional Conduct \citep{cfa_code}. 
    The policy covers professional integrity, independence and objectivity, misrepresentation, market manipulation, duties to clients, fair dealing, suitability, confidentiality, duties to employers, investment diligence, communication, record retention, conflicts of interest, and proper use of CFA-related designations. 

\end{itemize}

\section{More Experimental Results}
\label{app:more_results}

To further examine the effectiveness of PolicyAlign on a more recent backbone, we conduct additional experiments on Qwen3-4B.
The results in Table~\ref{tab:qwen3_4b} show that PolicyAlign achieves the strongest overall safety-utility trade-off on Qwen3-4B.
It achieves the lowest ASR on Advbench, WildJailbreak, and Fortress.
Compared with SFT, PolicyAlign reduces Fortress ASR from 17.0 to 13.6 and improves XSTest compliance from 67.2 to 87.6, suggesting that the safety gains do not simply come from more aggressive refusal.
Meanwhile, PolicyAlign preserves general capabilities, outperforming SFT on all general capability benchmarks.

\section{Broader Impacts}
\label{app:broader_impacts}

PolicyAlign aims to make LLM safety alignment more responsive to explicit safety policies, such as usage guidelines, professional norms, and domain-specific regulations. This may reduce the cost and latency of adapting models to emerging safety requirements, especially in high-stakes domains where authoritative policies are available but large-scale supervision data is scarce. By improving safety while preserving benign-query compliance, the framework may help reduce compliance with harmful requests and unnecessary refusals.

At the same time, policy-based alignment should be used with care. The behavior learned by the model depends on the quality, authority, and interpretation of the underlying policy. If a policy is incomplete, outdated, or biased, these issues may be reflected in the aligned model. In addition, the data generation process may produce policy-violating instructions, which could pose misuse risks if released without safeguards. Therefore, practical deployment should involve authoritative policy sources, human oversight, red-teaming, and periodic re-evaluation. For responsible release, generated data containing harmful instructions should be accompanied by misuse warnings, appropriate access controls or redaction of highly actionable details, and clear documentation of the policy sources.

\end{document}